\documentclass[letterpaper, 10 pt, conference]{ieeeconf}  

\IEEEoverridecommandlockouts                              

\usepackage{graphicx} 
\usepackage{amsmath} 
\usepackage{amssymb}  
\usepackage{url}
\usepackage{color}
\usepackage{hyperref}

\title{\LARGE \bf
Predicting Road Surface Anomalies\\ by Visual Tracking of a Preceding Vehicle
}

\author{Petr Jahoda, Jan {\v C}ech\\
Faculty of Electrical Engineering, Czech Technical University in Prague
}

\begin{document}

\maketitle
\thispagestyle{empty}
\pagestyle{empty}

\begin{abstract}
A novel approach to detect road surface anomalies by visual tracking of a preceding vehicle is proposed. The method is versatile, predicting any kind of road anomalies, such as potholes, bumps, debris, etc., unlike direct observation methods that rely on training visual detectors of those cases. The method operates in low visibility conditions or in dense traffic where the anomaly is occluded by a preceding vehicle.  
%
Anomalies are detected predictively, i.e., before a vehicle encounters them, which allows to pre-configure low-level vehicle systems (such as chassis) or to plan an avoidance maneuver in case of autonomous driving. 
A challenge is that the signal coming from camera-based tracking of a preceding vehicle may be weak and disturbed by camera ego motion due to vibrations affecting the ego vehicle. Therefore, we propose an efficient method to compensate camera pitch rotation by an iterative robust estimator. 
Our experiments on both controlled setup and normal traffic conditions show that road anomalies can be detected reliably at a distance even in challenging cases where the ego vehicle traverses imperfect road surfaces. The method is effective and performs in real time on standard consumer hardware.

%
%


\end{abstract}

\section{Introduction} \label{sec:intro}

Ensuring road safety and comfort is crucial in the contemporary transportation landscape. Identifying road surface anomalies, such as potholes, bumps, rutting, expansion joints, and debris, is important for the safeguarding of vehicle occupants, improving driving comfort, and reducing wear and tear on tires, wheels, or vehicle suspension components.  

Forward-looking systems that detect such irregularities before the vehicle encounters them are particularly important. These systems can issue timely alerts to human or autonomous drivers, or provide predictive information to enable appropriate responses of chassis subsystems, as e.g., suspension, traction, and braking.

Usually, camera-based predictive systems \emph{directly} recognize surface anomalies with clear visibility in front of a vehicle. Detecting road irregularities is not reliable in poor visibility conditions, due to weather conditions, the time of day, or often due to obstruction by other vehicles in traffic. 

\begin{figure}
    \centering
    \includegraphics[width=1\linewidth]{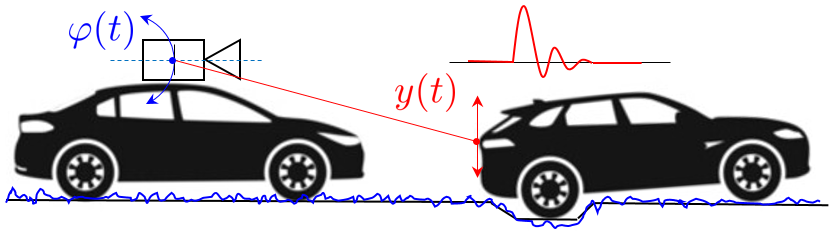}
    \caption{The main idea is to track the motion of a preceding vehicle using a camera mounted in the ego vehicle to detect anomalies on the road surface, such as potholes, bumps, etc. The vertical trajectory $y(t)$ measured by the camera reflects the surface profile beneath the preceding vehicle. Anomalies in this signal, are anomalies of the road surface. A challenge is that the camera of the ego vehicle moves. The translation motion is not important, but the camera pitch rotation $\varphi(t)$, caused by other uneveness of the road, distorts the signal. We visually estimate the ego rotation of the camera $\varphi(t)$ and compensate for it, which results in improved accuracy of the method.}
    \label{fig:motivation}
\end{figure}

This paper introduces a system that \emph{indirectly} detects road surface irregularities by monitoring the movement of a vehicle ahead. Our method automatically detects typical motion patterns indicative of surface anomalies, much like an experienced driver being able to recognize a problem by observing sudden jolts of the preceding vehicle, even without directly seeing the cause on the road. See Fig.~\ref{fig:motivation} for an illustration of the proposed approach.

The method is robust to visibility challenges, as the car rears are intentionally designed to be visible in any time/weather condition. Consequently, this method can provide early detection of common anomalies before the vehicle gets close enough to detect them directly.

We track a car rear in the image of a front-looking camera over time~\cite{Jahoda-robovis-2025}. A signal indicating that the preceding vehicle is going over a road surface anomaly may be weak and difficult to recover. Various adverse effects disturb the signal. In particular, a distance between vehicles and vibrations of the ego vehicle's own movement due to road imperfections. The signal is weaker with longer distances, and vibrations cause background noise that contaminates the tracking signal and decrease the sensitivity of the method. 

In this paper, we propose a technique to compensate for the camera ego motion, namely the pitch rotation, which is the dominant source of disturbances arising from the traversal vibrations. Unlike the traditional approach, which estimates the essential matrix between frames via RANSAC~\cite{nister2004efficient}, we propose an iterative procedure using a robust loss function applied to visual correspondences found by the state-of-the-art long-term optical flow~\cite{Neoral2024mft}. 

By compensating for the ego-camera rotation, we increase the detection accuracy of the method. We quantitatively show that this method outperforms the baseline introduced by~\cite{Jahoda-robovis-2025} and operates in challenging conditions on imperfect roads. The method is more accurate, with significantly fewer false positive detections, which we quantitatively show in several experiments.

\section{Related Work} \label{sec:sota}

The detection of anomalies on road surfaces has been addressed in the literature using multiple approaches. There are two main principles: (1) Effect-based methods, where a road surface defect is identified ex post by traversing, usually detected by accelerometers~\cite{martinez2022review}. (2) Predictive methods, where anomalies are spotted a priori before a vehicle encounters them, typically using a camera~\cite{jakubec2023comparison} or a lidar~\cite{talha2024use}. 
These methods include detection of potholes~\cite{kim2022review}, speed bumps~\cite{aishwarya2023robust}, or general pavement distress~\cite{du2021application}. For a more detailed taxonomy of these methods, we refer to a recent survey~\cite{rathee2023automated}. 

Recent vision-based methods are based on machine learning and deep models, utilizing supervised training with datasets of manually annotated images~\cite{kulambayev2023real}. An alternative approach to manual labeling is self-supervised learning, where a training label for images is supplied by an accelerometer, and then image-only model is used for deployment~\cite{cech2021self}. Another approach is to treat anomalies as unpredictable off-distribution samples~\cite{vojivr2023image}.

An example of a predictive commercial system is ``Magic Body Control'' by Mercedes-Benz~\cite{MercedesMBC}, which relies on a forward-looking stereo camera to monitor road-surface disturbances as, e.g., speed bumps, and communicates with the suspension controller to adjust real-time response to mitigate effects on the vehicle body.

Cooperative techniques to collect and share road-surface hazards have been considered for a fleet of vehicles~\cite{LandRover-potholes} or in a vehicle-to-vehicle scenario~\cite{zhu2023road,stone2019connected,rahman2012using}. 

Our work aims to predictively detect road surface anomalies using the perception system of the ego vehicle, which indirectly infers anomalies from the motion pattern of the preceding vehicle using a camera. This system is complementary to direct sensing methods and can be advantageous in adverse conditions, such as poor visibility, or where the preceding vehicle obscures a road anomaly, which is frequent in urban traffic. 

A close work to our approach is~\cite{li2022estimating} which adopts a control engineering perspective. It focuses on solving an inverse problem of estimating the road profile by modeling the vehicle's vertical dynamics and employing a Kalman filter. However, their experiment does not demonstrate detection of anomalies from the perspective of an ego vehicle, but uses a static camera observing several vehicles from the side as they drive over a single pothole, which is not our target scenario.

Connecting the baseline method introduced in~\cite{Jahoda-robovis-2025} with camera ego-motion compensation is related to the digital video stabilization problem~\cite{wang2023video}. However, these methods target shakiness of the captured video by removing the high frequency component of global motion to make the resulting video more visually pleasant for a human. Our target is different; a global relative pitch rotation is estimated, and the estimate is used to compensate the tracked trajectory. 

\section{Method} \label{sec:mathod}

\begin{figure*}[t]
\begin{center}
\includegraphics[width=1.0\linewidth]{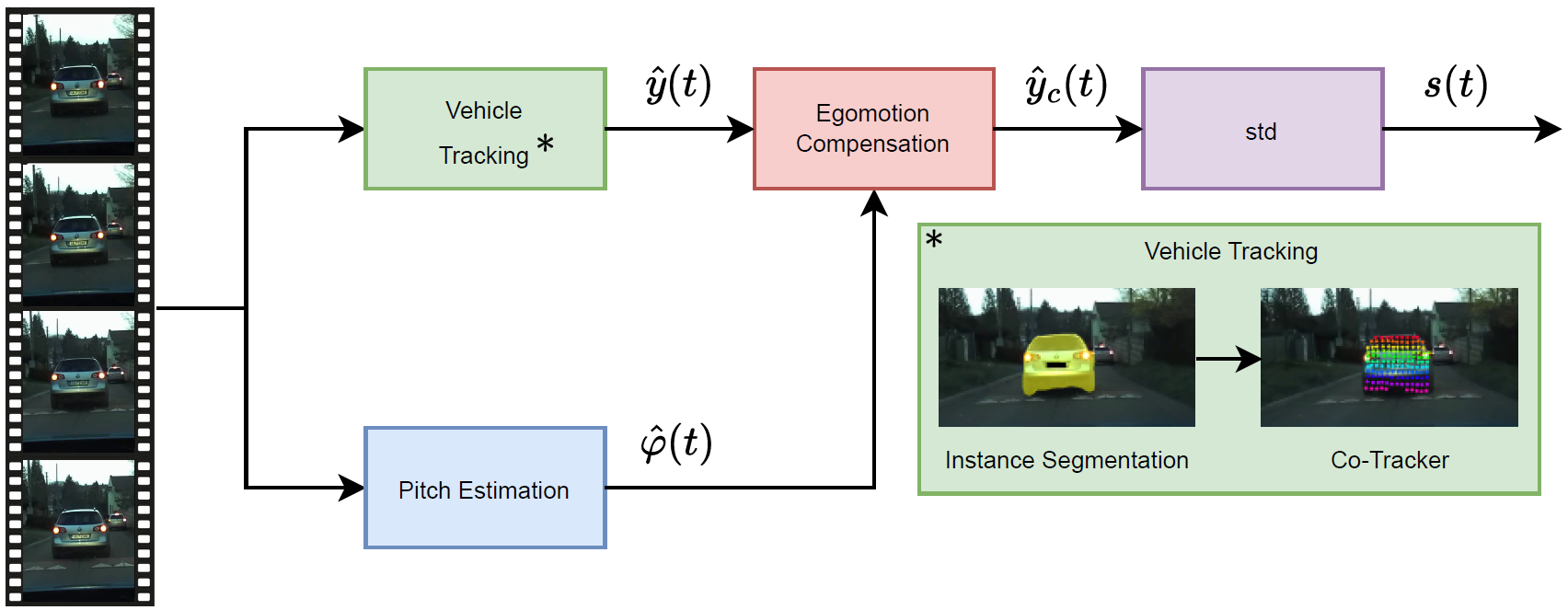}
\end{center}
   \caption{Flowchart of the proposed method. The vertical trajectory of the preceding vehicle is tracked by instance segmentation and CoTracker~\cite{karaev2023cotracker}. The pitch is estimated from correspondences on static background features, and the egomotion is compensated. Finally, the resulting visual response to the road profile is calculated as standard deviation of the compensated trajectory in a short window~(\ref{eq:std}).}
\label{fig:method}
\end{figure*}

The proposed method detects a preceding vehicle in the image of a front-looking camera and visually tracks it over time. Surface anomalies are found by detecting excessive deviations of the trajectory around the mean location. 

The pipeline of the algorithm is sketched in Fig.~\ref{fig:method}. It consists of the preceding vehicle tracking, which provides a vertical trajectory, pitch rotation estimation used to compensate for the ego motion of the camera, and finally the trajectory is aggregated to provide a visual response to a road surface profile. 

\subsection{Vehicle tracking and visual response calculation}
First, all vehicles are found by instance segmentation. We use Mask2Former~\cite{cheng2021mask2former} which delivers instance segmentation masks for all vehicles in the frame.

The target vehicle is found to be the closest to the center of the image. Then the vehicle is tracked by CoTracker~\cite{karaev2023cotracker}, which jointly tracks a grid of points. We set $N$ points equidistantly distributed over the instance mask. The tracker returns respective horizontal and vertical locations $x_i(t), y_i(t)$ for every grid point $i$ in frame $t$. Since we are only interested in vertical motion, we aggregated over vertical locations of tracked points $\hat{y}(t) = \frac{1}{N} \sum_{i=1}^N y_i(t)$.  Next, the signal indicating excessive displacement is calculated as standard deviation in a short window of $T$ frames
\begin{equation}
    s(t) = \sqrt{\frac{1}{T} \sum_{\tau = t-T+1}^t \bigl( \hat{y}(\tau) - \bar y \bigr)^2}, \label{eq:std}
\end{equation}
where $\bar y =  \frac{1}{T} \sum_{\tau= t-T+1}^t \hat{y}(\tau)$ is the mean location. Signal $s(t)$ is finally thresholded (with non-maximum suppression) to detect anomalies on the road surface and to distinguish them from a normal background.

\subsection{Visual pitch angle estimation and compensation} \label{pitchestimation}
 {The pitch rotation due to camera ego motion distorts the above signal and needs to be compensated. The pitch is estimated from corresponding features between frames. The correspondences are found by MFT Tracker~\cite{Neoral2024mft} computed at dense grid points and good features to track~\cite{lucas1981iterative}. To avoid contamination by moving parts of the scene, all points corresponding to cars are removed.} 
 

{Given the assumption that the movement is purely forward with negligible roll or yaw, we estimate the pitch angle from correspondences}
\begin{equation}
\hat{\varphi}(t) = \arg\min_{\varphi} \sum_{i} \rho \left( S(\mathbf{p}_i(t), \mathbf{p}_i(t+1), \mathbf{F}(\varphi)) \right),  
\label{eq:opt}
\end{equation}
where $\mathbf p$ are correspondences between subsequent frames $t$ and $t+1$,  $\mathbf F$ is the Fundamental matrix parameterized only by pitch angle $\varphi$, $S$ is a Sampson error~\cite{Hartley} p.~287, and $\rho$ is the Cauchy loss defined as $\rho(z) = \mbox{log}(1+z)$. 

The fundamental matrix~\cite{Hartley} p. 257 is parametrized as
\begin{equation}
     \mathbf{F}(\varphi) = - \mathbf{K}^{-T} [\mathbf{t}]_{\times} \mathbf{R}(\varphi)  \mathbf{K}^{-1},
\end{equation}
where $\mathbf{K}$ is the intrinsic camera matrix, $\mathbf t$ is the translation vector representing the forward direction of the car, $[\mathbf t]_\times$ stands for the skew-symmetric matrix of the vector, and $\mathbf R(\varphi)$ is the pitch rotation matrix. Both $\mathbf{K}$ and $\mathbf t$ are calibrated offline. 

The optimum of problem~(\ref{eq:opt}) is found by the Levenberg-Marquardt algorithm initialized from a previous frame solution. The first frame is initialized to zero. 

The compensation of the estimated trajectory is then applied as  
\begin{equation}
    \hat{y}_c(t) = \hat{y}(t) - f \cdot \tan \hat{\varphi}(t) , \label{eq:comp}
\end{equation} where $f$ is the focal length in pixels and $\hat{\varphi}(t)$ is the predicted pitch.


{In all experiments, we set the temporal window $T=30$ frames, which represent one second in our setup. Points tracked by CoTracker were equidistantly distributed over the instance masks, by 6 pixels vertically and horizontally by default. We limit the maximum number of points to $N = 400$ to fit in GPU memory.}

\medskip

 \paragraph*{{Discussion}} 
 {Note that anomalies are detected by thresholding the signal $s(t)$ in Eq.~(\ref{eq:std}), relying on tracking in the image plane (in pixels). In principle, it is possible to convert the displacements into the scene based on the vehicle distance and set the threshold in meters. However, accuracy of the displacement estimate in the scene obviously depends on the distance, which would need to be taken into account, see our analysis in Sec.~\ref{sec:signal_strength}. Otherwise, false detection would easily occur at a distance. This is not a problem when the thresholding is done in pixels, as the displacement in pixels is naturally small for a distant target.}

{The pitch angle to compensate for the camera ego rotation is estimated visually. It relies on trackable features in the scene, which is not usually a problem to find using modern trackers. In principle, the pitch of the camera could be estimated using an IMU without any visual assumptions. However, with IMU, it is necessary to integrate the angular velocity to estimate the rotation angle, and it tends to drift.}
 

A simplification we made is that we do not model the vehicle suspension. Clearly, the vehicle body motion response to the excitation coming from road anomalies is not linearly proportional to the depth or height of the road irregularity. It is a complex function dependent on many factors, such as axle and suspension kinematics, tuning of the vehicle suspension, vehicle body dynamics, etc. 
The simplification is justified because we target urban traffic situations, which typically occur at lower speeds where excitation signals exhibit dominant low-frequency spectra, making anomalies clearly detectable in vehicle body motions. We frame the problem as a binary classification: the positive class represents surface anomalies, while the negative class corresponds to the natural background. 

\section{Datasets} \label{dataset}
We collected a dataset called SVAR (Sources of Vertical Acceleration on the Road), consisting of two subsets. The first subset is acquired for a controlled experiment, using two instrumented cars and well-defined road surface anomalies and traversing scenarios. The second subset was acquired in normal traffic, where the ego car was lightly instrumented with a camera and IMU, usually of a smartphone. 

        

\begin{figure}
    \def \x {0.24}
    \hspace{-7mm}\footnotesize{Controlled} \hspace{2.55cm} \footnotesize{In-the-wild subset} \\[1ex]
    \centering
        \includegraphics[width=0.22\linewidth]{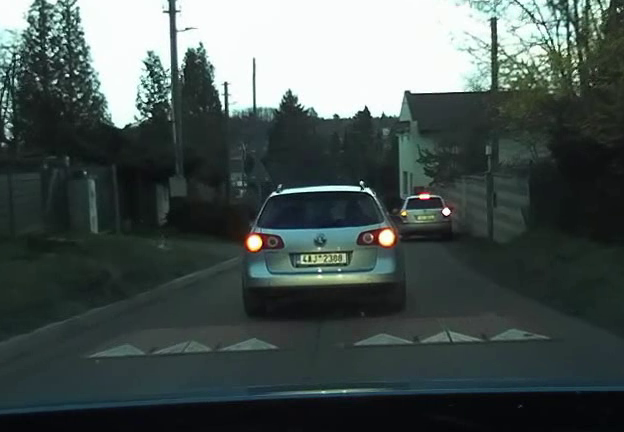} \hspace{2pt}
        \includegraphics[width=\x\linewidth]{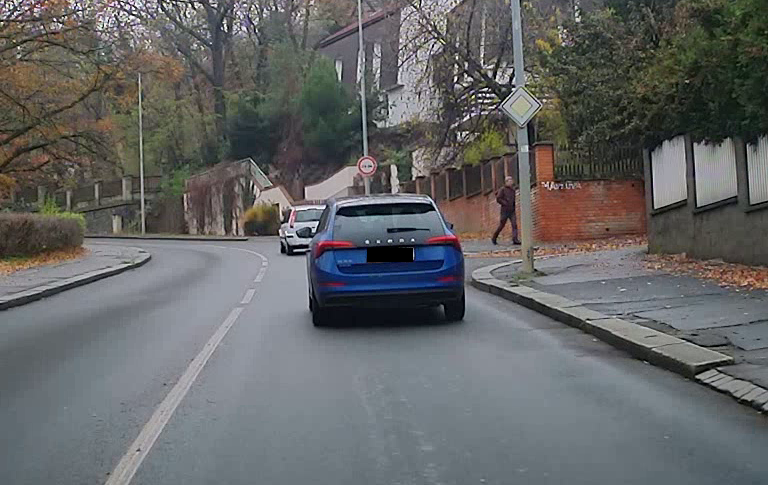}  
        \includegraphics[width=\x\linewidth]{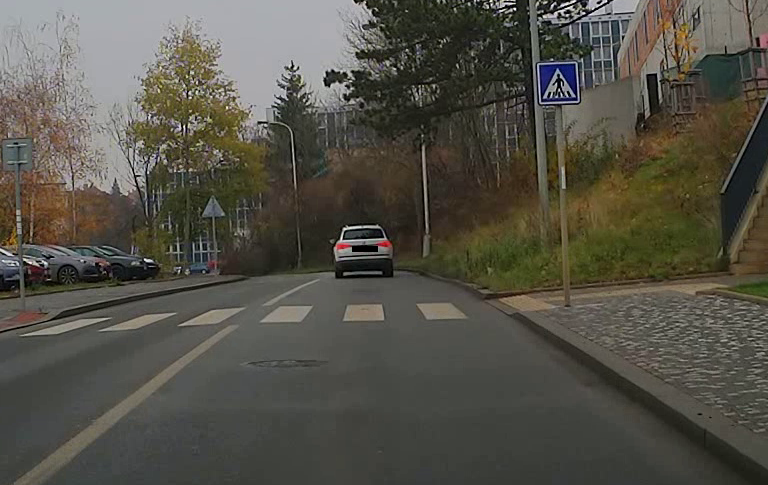} 
        \includegraphics[width=\x\linewidth]{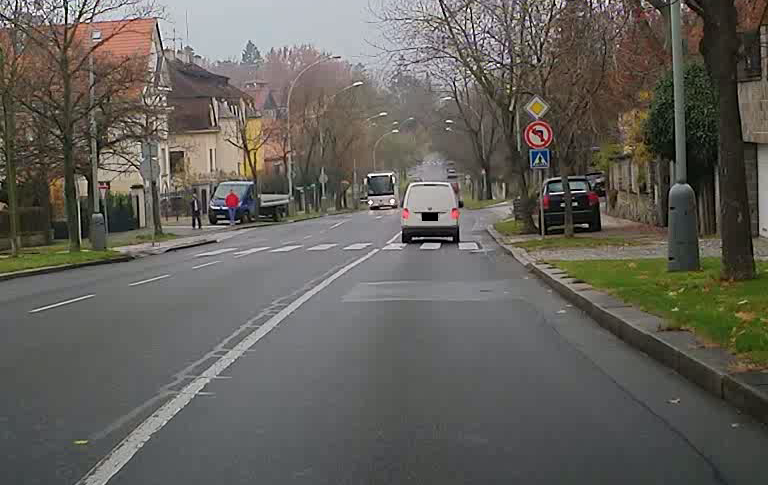} 
    \\
    \vspace*{1mm}
    \centering
        \includegraphics[width=0.22\linewidth]{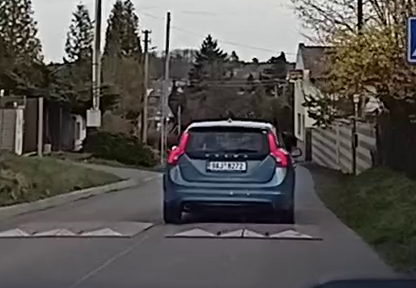} \hspace{2pt} 
        \includegraphics[width=\x\linewidth]{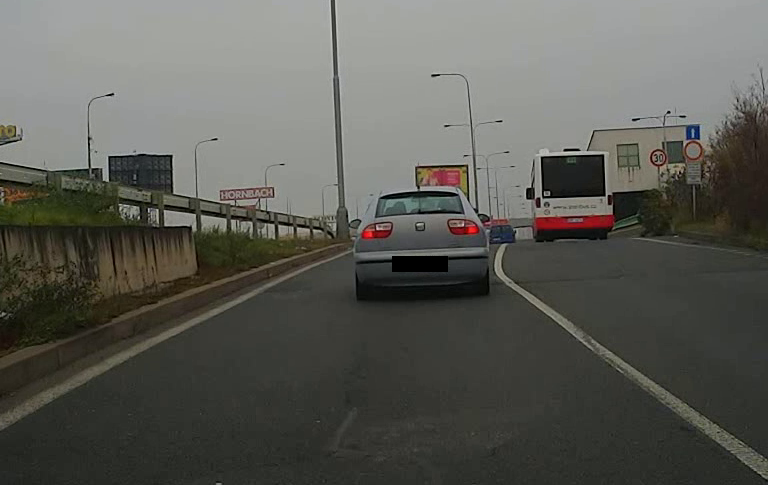}  
        \includegraphics[width=\x\linewidth]{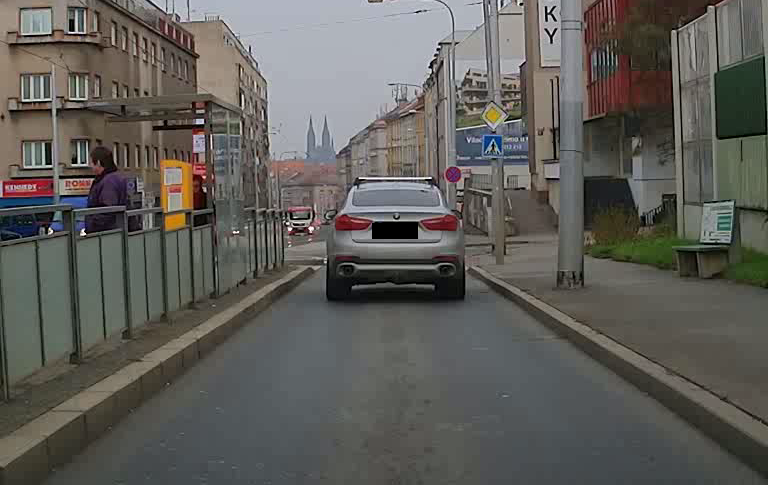} 
        \includegraphics[width=\x\linewidth]{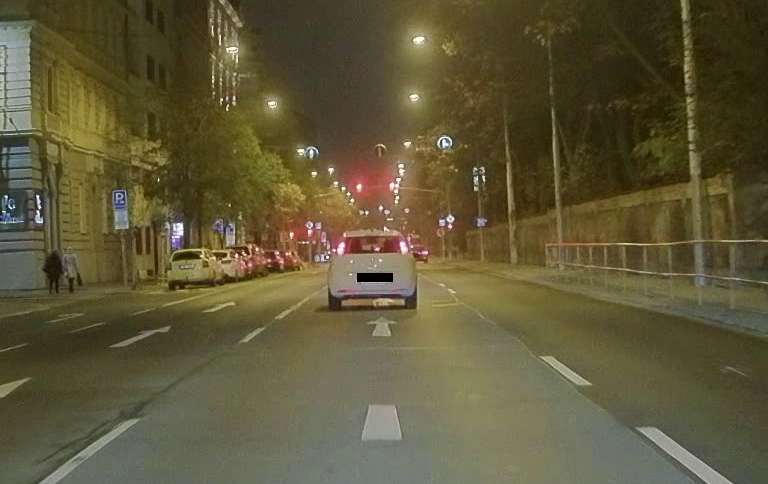}  
    \caption{Examples of anomalies, sources of high-vertical acceleration, on the road surface in our SVAR dataset, controlled and int-the-wild subsets. Note that except for the leftmost column, the road surface anomalies are not visible to a human observer, but are detectable by observing the preceding vehicle. For better illustration, see supplementary videos at 
    \protect\url{https://cmp.felk.cvut.cz/\%7Ecechj/video/iv-2025/} .}
    \label{fig:dataset}
\end{figure}

\subsubsection{Dataset for controlled experiment} \label{sec:controlled_dataset}
We used two cars that were instrumented in the same way. In particular, ZED2 camera was mounted behind the windscreen, and IMU was attached to the floor of the trunk. Finally, an automotive-grade GPS (with an external antenna on the roof) was installed. Each instrument was connected to the computer that recorded the data. The synchronization of the recordings between the cars was done using GPS time stamps. 

Two cars repeatedly crossed three same speed bumps (of dimensions $2\times1.78\times 0.06$ meters, both approaching and exiting edge of $7^\circ$, see left most column in Fig.~\ref{fig:dataset}) in various mutual distances from about {3 to 40 meters}. The speed of the traversal was about 20 km/h. 

Synchronized GPS units give us precise measurements of the distance between the cars. The IMU in the preceding vehicle provides automatic annotation of the event when it travels over the speed bump.

\subsubsection{In-the-wild dataset}
We collected anomalies on the road surface in normal traffic conditions. In this case, a single car was instrumented with a camera and an accelerometer. The first part of the dataset was captured using consumer smartphones, the second part with a more advanced test vehicle, but only a standard automotive camera and an accelerometer were used. 

The annotation process was semi-automatic. When an excessive vertical acceleration was found, we checked whether a car detector~\cite{cheng2021mask2former} spots a vehicle in front. If so, we manually inspected whether the preceding vehicle went over the same surface anomaly and annotated the apex frame of the event. 

In total, we collected 157 events when a preceding car encountered road surface anomalies. {In 92 of the events, the ego vehicle simultaneously goes over other road surface anomaly too, which we refer to as a hard subset.} Short video clips were cut out of long recordings, 5 seconds prior and 5 seconds post the apex frame. The anomalies are of various nature, e.g., speed bumps, potholes, or other road depressions, unaligned surface levels when road segments connect, or ``wavy'' road surfaces. Sometimes the anomalies are hardly or not at all visible directly, and the only indication of the problem is the motion pattern of the preceding vehicle. {Several examples are shown in Fig.~\ref{fig:dataset} and in our supplementary videos at \url{https://cmp.felk.cvut.cz/\%7Ecechj/video/iv-2025/}.}

{These events are relatively rare. In particular, 157 events were found in about 15 hours-long recordings. The occurrence depends on the road quality and traffic density, since a preceding vehicle indicating the anomaly is needed. We got most of the events for urban driving in rush hours and very few for rural roads.}



\section{Experiments}

\subsection{Signal strength as a function of the vehicle distance}
\label{sec:signal_strength}
\begin{figure}[t]
    \centering
    \includegraphics[width=0.95\linewidth]{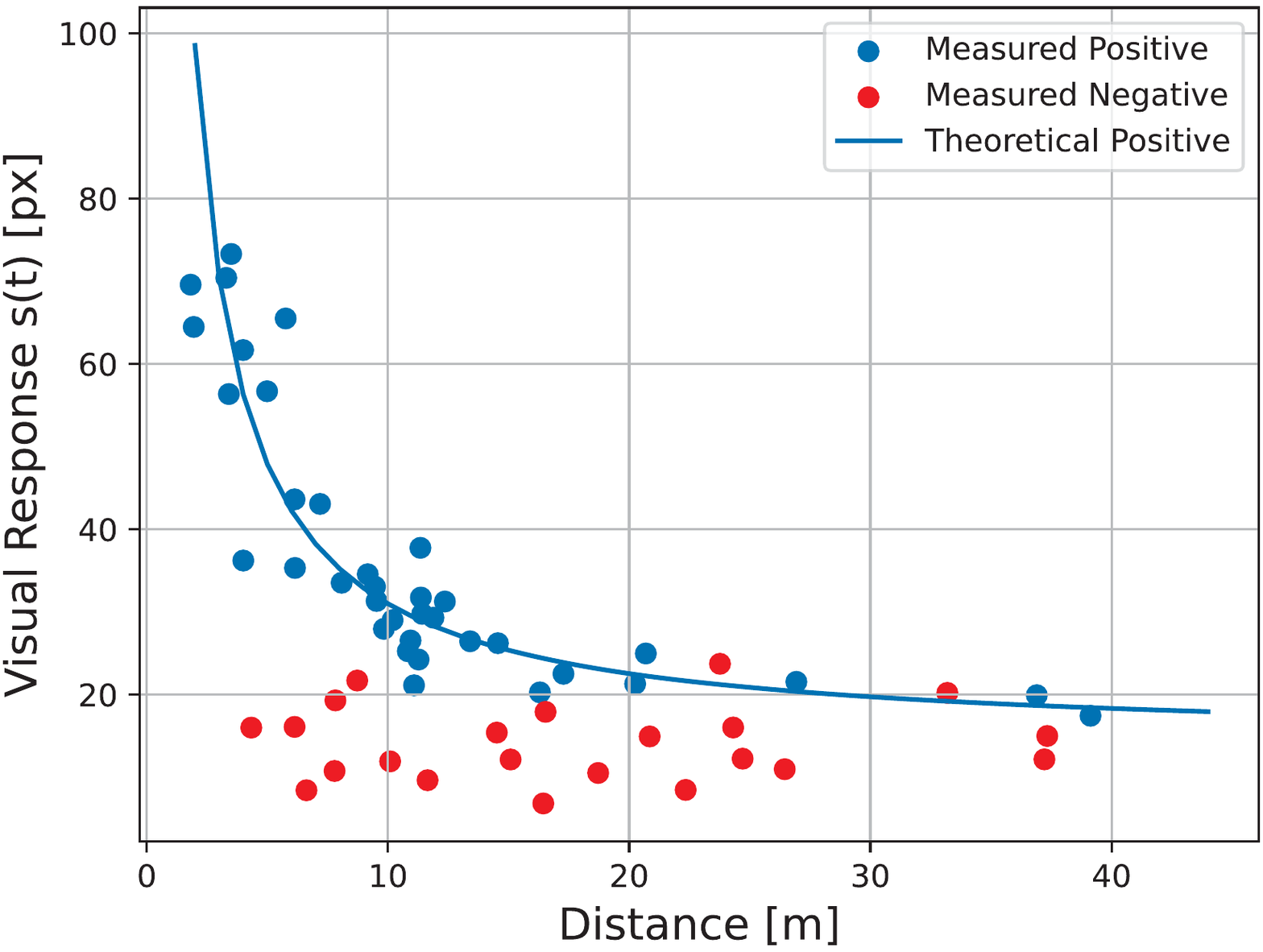}
    \caption{Visual responses $s(t)$ estimated by Eq.~(\ref{eq:std}) vs. car distances for the controlled experiment.  Plot shows positive class samples (speed bumps), negative class samples (natural background), and the theoretical expectation given by our model in Eq.~(\ref{eq:theory}).}
    \label{fig:controlled_exp}
\end{figure}

In the following experiment, we measured, on the controlled dataset described in Sec.~\ref{sec:controlled_dataset}, how the strength of the signal depends on the distance to the preceding car. In Fig.~\ref{fig:controlled_exp}, we show the visual response $s(t)$ in Eq.~(1), computed at the time of traversal over the speed bump for various distances. Obviously, the response is stronger the closer the two vehicles are. 

Let us assume that due to a speed bump, a vehicle moves in the scene vertically by displacement $\delta$ (in meters). In theory, observing the vehicle from behind by a perspective fronto-parallel camera, with focal length $f$ (in pixels) in distance $d$ (in meters), then the displacement in the image plane (in pixels) is given by $\Delta y =f \frac{\delta}{d}$. Theoretical visual response $s(t)$, which we measure in Eq.~(\ref{eq:std}), is the sum of two processes -- the excitation due to the anomaly and due to the natural background. That is
\begin{equation}
   \hat s(d) = \alpha f \frac{\delta}{d} + \beta. \label{eq:theory}
\end{equation}
The camera captures Full-HD $1920\times1080$ image, vertical field of view is $54^\circ$, which means $f = 1066$ pixels. {We set $\delta = 0.06$ m, a height of the speed bump.} Parameters $\alpha$ and $\beta$ are fit in the least square sense.

Figure~\ref{fig:controlled_exp} shows the visual responses corresponding to the positive class (speed bumps) and its theoretical expectation given by the above equation. Negative class samples, i.e. visual responses of random locations off the speed bumps (and other anomalies), correspond to a natural background due to fluctuations of the tracker or due to ego vehicle motion caused by imperfections of the road. The visual response is seen to be above the background level if the distance is less than {20 meters}. 

\subsection{Evaluation of camera pitch-rotation and compensation}
\begin{figure}[t]
    \centering
    \includegraphics[width=0.83\linewidth]{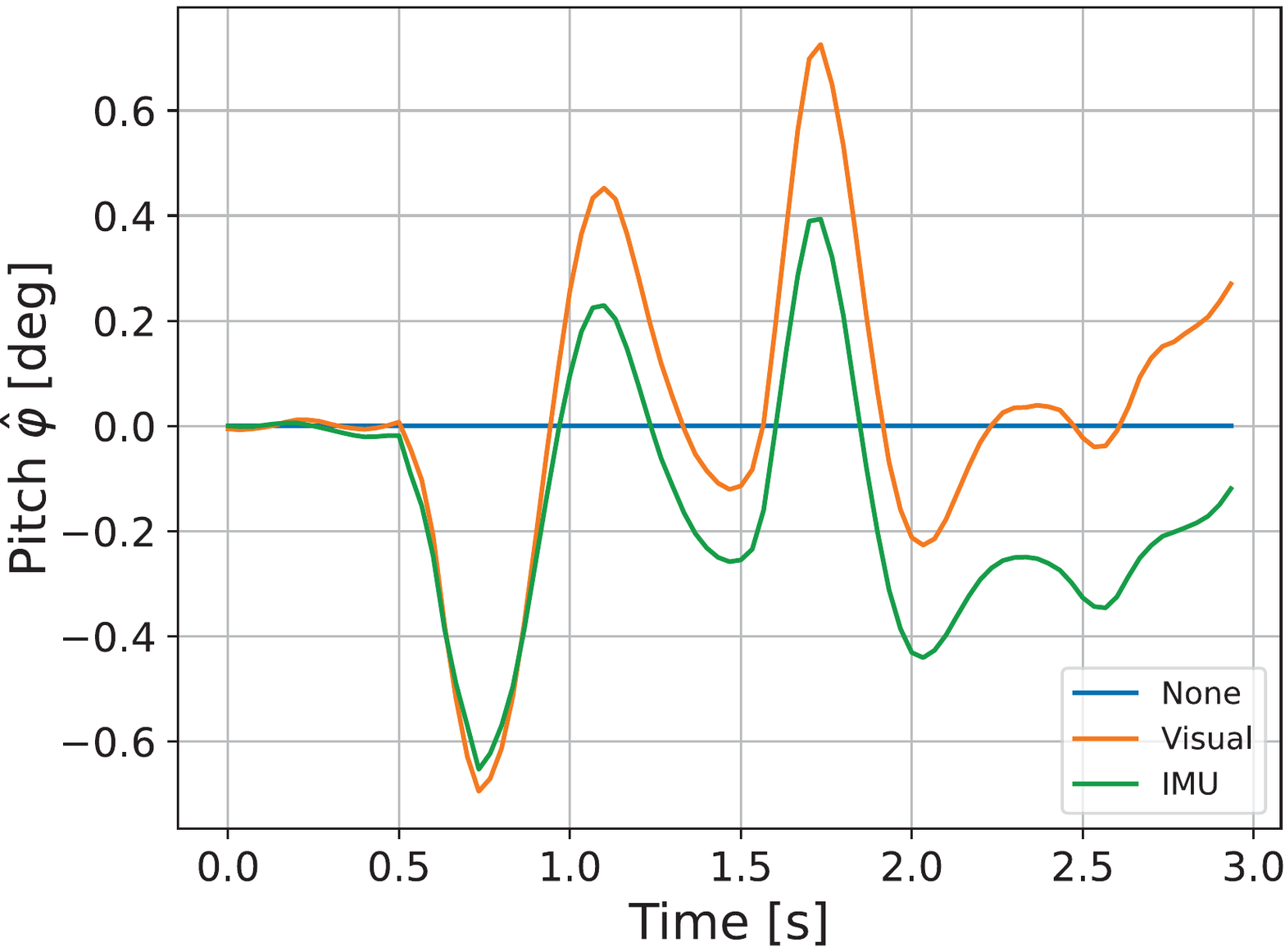} \\[3ex]
        \includegraphics[width=0.83\linewidth]{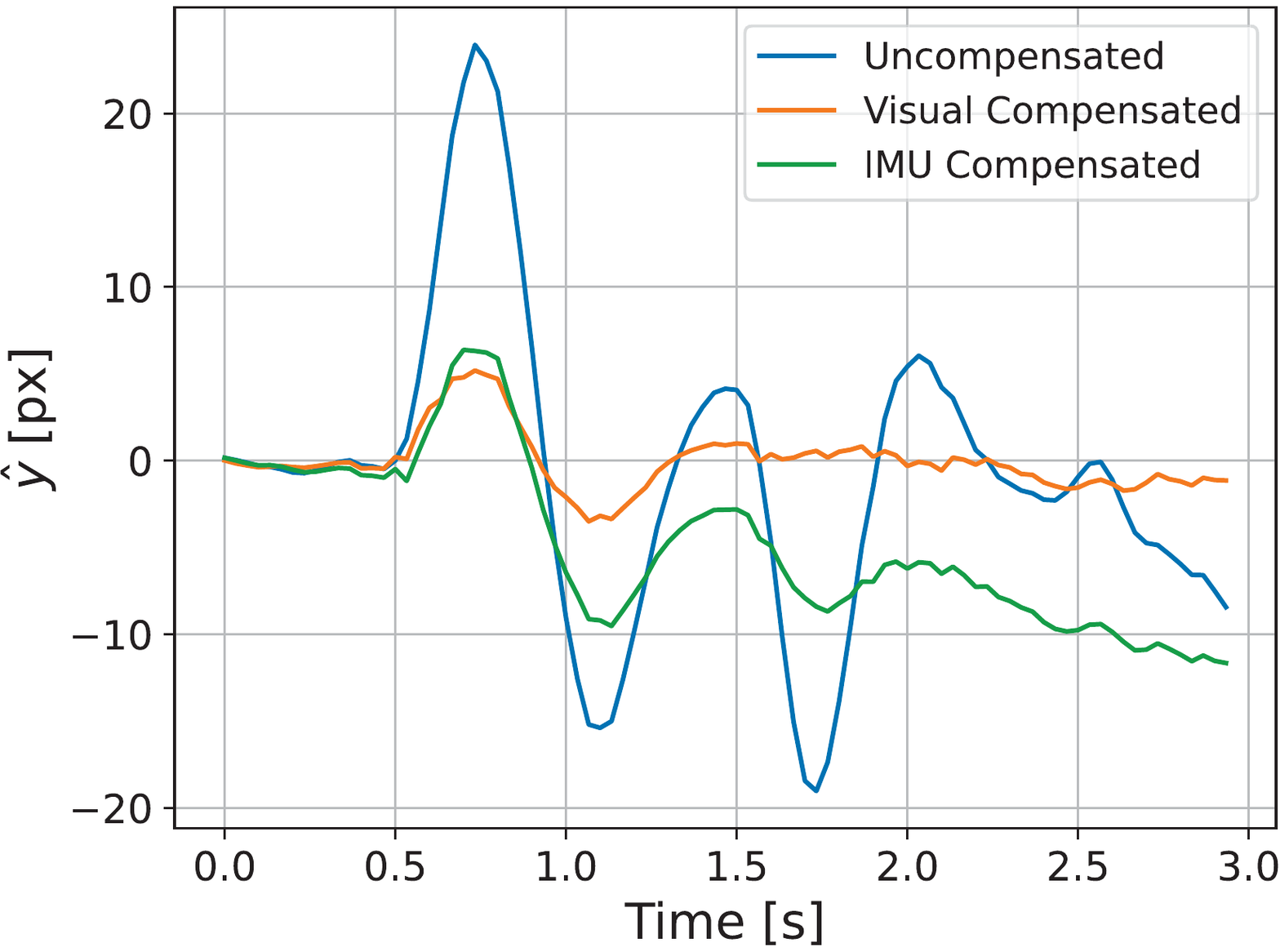}
    \caption{Prediction of pitch angle of an ego vehicle traversing a speed bump (top). Corresponding trajectory of the preceding car with the egomotion compensated using the estimated pitch (bottom).}
    \label{fig:Angles}
\end{figure}

Following experiments are conducted on the controlled dataset. In Fig.~\ref{fig:Angles} (top), we show an example of camera pitch angle, when the ego vehicle traverses a speed bump, estimated both visually and by IMU (of a smartphone). The estimated angle is used for compensation of the ego-camera rotation when tracking the preceding vehicle. At this instance, the preceding vehicle does not go over any uneven surface. The estimated vertical trajectory $y(t)$ should be close to zero in Fig.~\ref{fig:Angles} (bottom). Three cases are plotted: without compensation and with pitch estimated visually and by the IMU. The visual compensation works the best. The IMU provides an angular velocity and is integrated, which causes a drift, i.e. an imprecise estimate of the pitch angle with time, which results in poor compensation. 

\begin{figure}[t]
    \centering
    \includegraphics[width=0.85\linewidth]{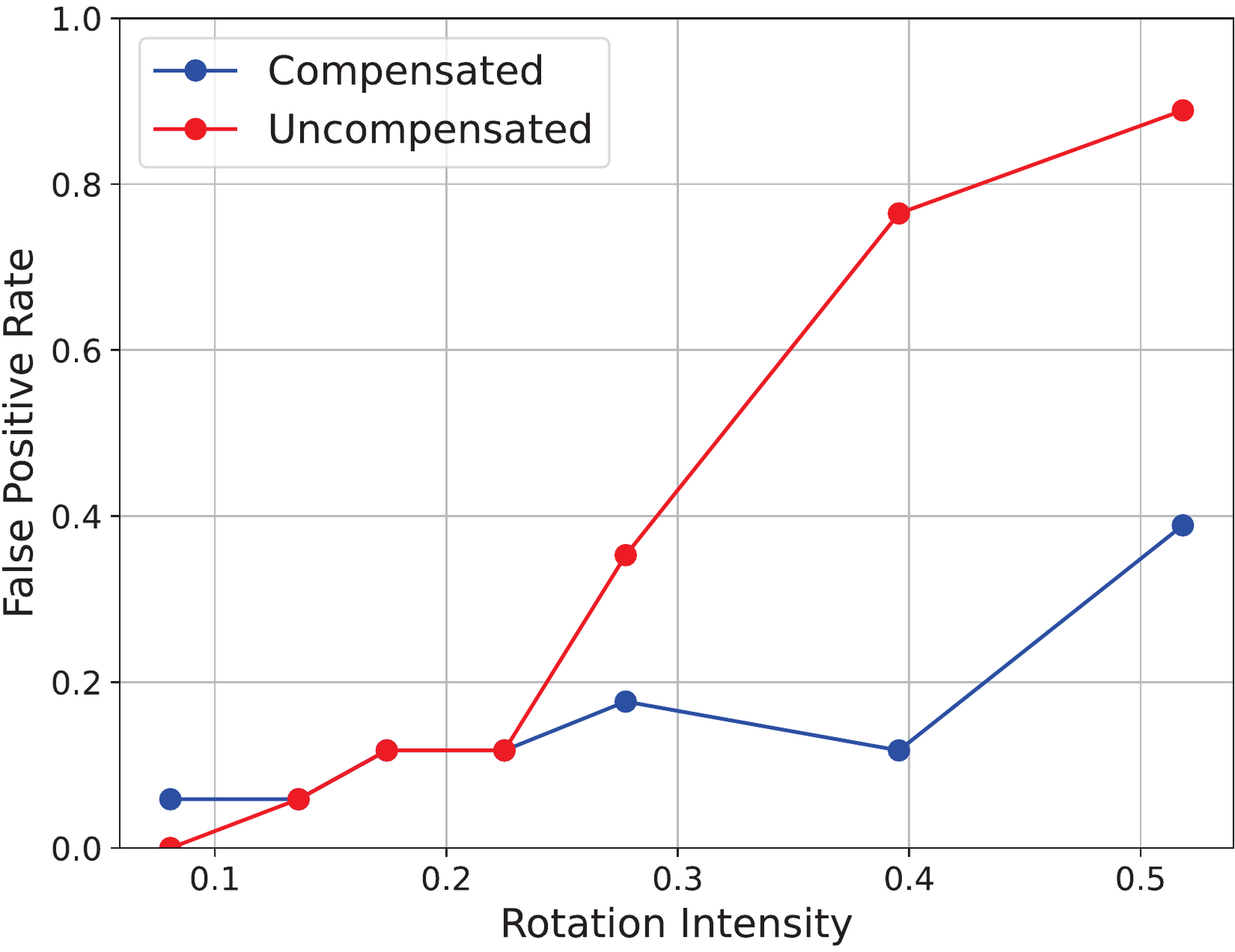}
    \caption{False Positive Rate (FPR) curves without and with the proposed pitch rotation compensation. Rotation intensity is calculated as averege absolute pitch angle over 1-second window.}
    \label{fig:Integral}
\end{figure}

Uncompensated camera ego rotation causes false positive detections of road anomalies. We evaluated the effect and benefits of the compensation in Fig.~\ref{fig:Integral}. The ego vehicle traverses over speed bumps, while the preceding vehicle is going over a smooth surface. False Positive Rate (FPR) is measured as a function of the ego rotation intensity, which is an average absolute angle in a 1-second window. It is seen that FPR is significantly reduced by the compensation, and the difference is increased with higher ego rotation intensities. 

\subsection{Accurcy of the road surface anomaly detector}
In this section, we evaluate the accuracy of our proposed road surface anomaly detector. 

The following results are calculated on the challenging in-the-wild dataset. We added about the same number of random background events to the dataset of road surface anomalies, in order to have samples of both positive and negative classes. 

We compare the proposed method integrating our pitch rotation estimation with two baselines. The first baseline does not compensate and the second uses an estimate of the essential matrix between frames~\cite{Hartley}, which is a standard approach to estimate the relative motion of the camera. The essential matrix is robustly estimated using the 5-point RANSAC algorithm from correspondences~\cite{nister2004efficient}. The pitch rotation is then extracted from the essential matrix. 

The quantitative comparison is shown in Table~\ref{tab:Quantitative_Evaluation}. To compare Balanced Accuracy\footnote{Balanced Accuracy is an average of accuracies on positive and negative classes.} and F-score, the threshold of visual response in Eq.~(\ref{eq:std}) that separates the positive (road surface anomaly) and negative (background) classes is {set to maximize the F-score for each one of the methods}. {This is done in a cross-validation setup. The optimum threshold is estimated on the training set and tested on the unseen subset. The results are averaged and standard deviation over the five-fold cross-validation is shown in Tab.~\ref{tab:Quantitative_Evaluation}}. AUC is the area under the ROC curve. 

It is seen that the full method using our pitch compensation performs the best on the entire dataset and on both hard and easy splits. The easy split contains only cases where the ego vehicle does not go over any significant anomaly and no pitch compensation is involved. The compensation using the essential matrix is less accurate, since it estimates many degrees of freedom of the motion and does not use the property of continuous motion in a video, as it performs the estimate in every frame from scratch. Moreover, the RANSAC-based estimation of the essential matrix is much slower than our proposed method. 

\begin{figure}[t]
    ~\\[1ex]
    \centering
    \includegraphics[width=0.85\linewidth]{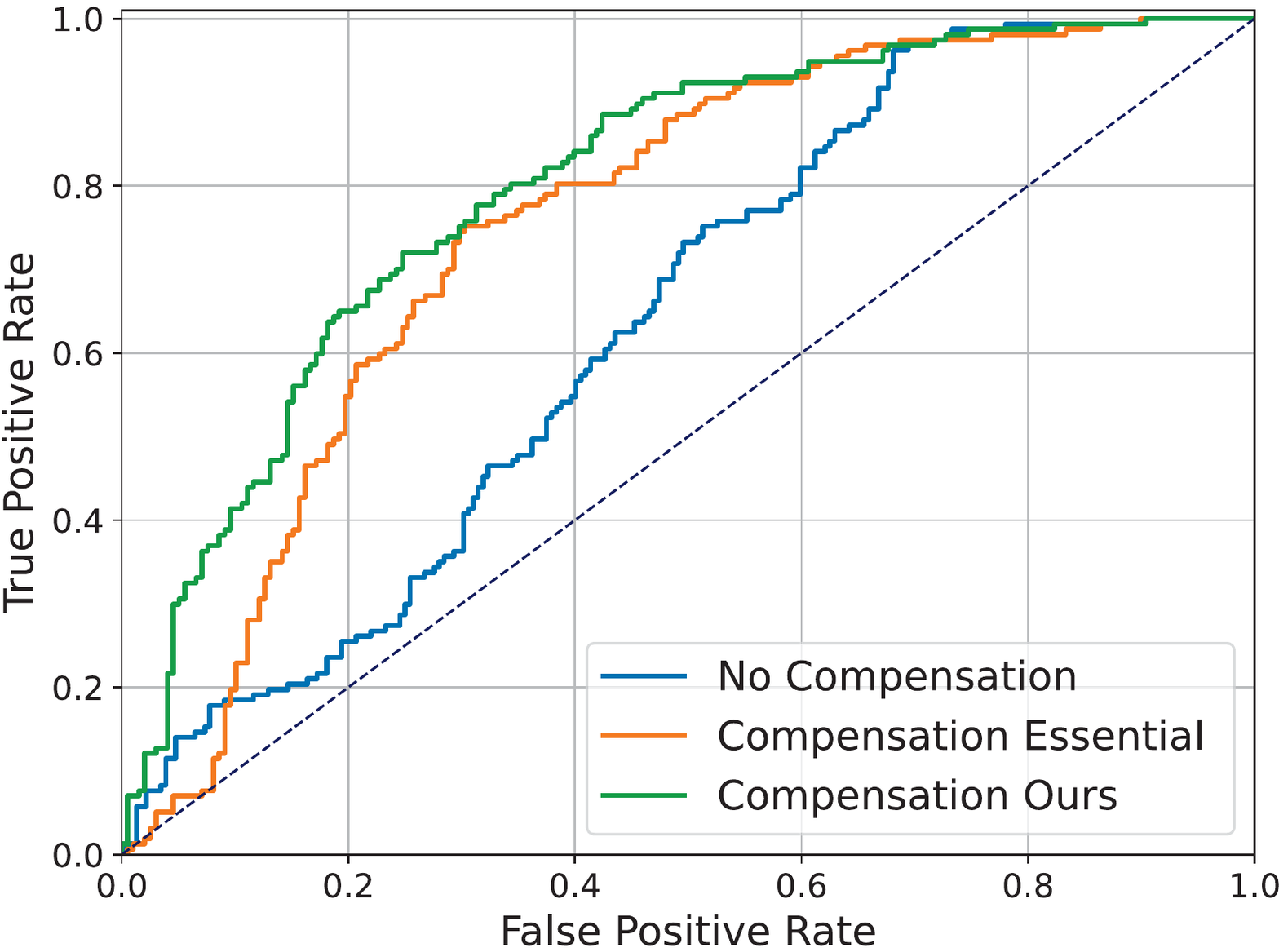}
    \caption{ROC curves for the tested methods.}
    \label{fig:ROC}
\end{figure}

To provide further insight, we show ROC curve in Fig.~\ref{fig:ROC}. The method with our proposed compensation outperforms other baselines for all thresholds. It has a higher or equal true positive rate and a lower false negative rate than other baselines. 


\begin{table}[t]
\centering
\caption{Quantitative comparison of the Methods. }
\resizebox{0.9\linewidth}{!}{\begin{tabular}{|c|ccc|}
 \hline
 Entire Dataset& Bal. Accuracy$\uparrow$ & F-score $\uparrow$ & AUC$\uparrow$\\
 \hline
 Uncompensated & $0.630 \pm 0.014$&  $0.641 \pm 0.026 $&  $0.637$ \\
 Essential matrix  & $0.674 \pm 0.045 $& $0.674 \pm 0.045$ &  $0.753$  \\
 Ours   & $\mathbf{0.712 \pm 0.045} $& $\mathbf{0.712 \pm 0.045}$ & $\mathbf{0.801}$ \\
 \hline
 \hline
 Hard Subset & & & \\
 \hline
 Uncompensated & $0.506 \pm 0.013$&  $0.527 \pm 0.052 $ &  $0.529$ \\
 Essential matrix   & $0.525 \pm 0.057 $& $0.585 \pm 0.071$ &   $0.588$  \\
 Ours   & $\mathbf{0.628 \pm 0.007} $& $\mathbf{0.648 \pm 0.045}$ & $\mathbf{0.662}$ \\
 \hline
 \hline
 Easy Subset & & &\\
 
 \hline
 Ours   &$\mathbf{0.915 \pm 0.05}$  & $\mathbf{0.903 \pm 0.06}$ & $\mathbf{0.969}$ \\

 \hline
\end{tabular}}
\vspace*{2mm}
\label{tab:Quantitative_Evaluation}
\end{table}

\section{Conclusions} \label{sec:conclusion}

We introduced an innovative method for detecting any type of road surface irregularities that result in notable vertical acceleration, by monitoring the movement of vehicles ahead. We presented a method that efficiently compensates an ego rotation of the camera caused by an ego vehicle traversing uneven road surfaces or other anomalies simultaneously. We showed improved accuracy compared to cases where this effect is neglected. 

The high accuracy of our road surface anomaly detection system confirms that visual cues from the vehicle ahead can reliably indicate road conditions. Our SVAR datasets include examples in which drivers cannot see the anomaly on the road surface, but the system identifies it through the movement of the preceding vehicle.

The method is computationally efficient and operates in real-time on consumer-grade hardware.

A limitation of the current approach is that the visual pitch estimation needs a textured scene around the vehicle. This is usually not a problem, as many features can be tracked with modern trackers in typical environments. For extreme cases, e.g. a dense fog or a tunnel, visual pitch estimation might not work, and one has to rely on the IMU. Nevertheless, a failure of the visual estimator is easy to detect, and then the IMU reading can be used instead, or as a future work a robust estimate using fusion visual+IMU will be used. 


%



\bigskip
\noindent
{\bf Acknowledgment.} The research was supported by Toyota Motor Europe, and by CTU student grant under project SGS23/173/OHK3/3T/13.

\bibliographystyle{IEEEtranS}
\bibliography{bumps}

\end{document}